\documentclass[11pt]{article}

\usepackage[margin=1in]{geometry}
\usepackage{amsmath,amssymb}
\usepackage{newtxtext,newtxmath}
\usepackage{booktabs}
\usepackage{graphicx}
\usepackage{url}
\usepackage[hidelinks]{hyperref}

\begin{document}

\thispagestyle{empty}
\vspace*{0.5cm}

\noindent\rule{\textwidth}{2pt}

\vspace{0.45cm}
\begin{center}
  {\LARGE\bfseries Attention as a Routing Graph:\\[0.15em]
  Live Circuit Extraction from a Single Forward Pass}
\end{center}
\vspace{0.35cm}

\noindent\rule{\textwidth}{0.8pt}

\vspace{0.6cm}

\begin{center}
\begin{minipage}[t]{0.3\textwidth}
\centering
{\bfseries Ash Manvi}\par
Aquin Labs\par
{\ttfamily ash@aquin.app}
\end{minipage}
\hfill
\begin{minipage}[t]{0.3\textwidth}
\centering
{\bfseries Samreena Tajreen}\par
Aquin Labs\par
{\ttfamily samreena@aquin.app}
\end{minipage}
\end{center}

\vspace{0.55cm}

\begin{center}
  {\Large\bfseries Abstract}
\end{center}

\begin{quote}
Finding circuits in language models usually means running many careful
interventions. We try something simpler: treat attention as a routing map from
one forward pass, keep a small set of routes that point toward the answer, and
ask whether those routes actually matter.

They often do. On induction and IOI (tasks where the ``right'' circuit is
already known), ablating our extracted edges hurts the model much more than
ablating a random set of the same size. We evaluate \(n{=}100\) prompts per
cell on GPT-2 Small, GPT-2 Medium, and Pythia-410M, with paired gap tests and
bootstrap confidence intervals. The extract step costs one forward; a
head-by-head patch sweep costs about two orders of magnitude more.

We are not claiming a complete circuit atlas. We are claiming a cheap sketch
that carries real causal signal on known tasks, with clear failure modes when
it does not. Code and evaluation artifacts are at
\url{https://github.com/Aquinf03/live-circuit-routing}.
\end{quote}

\section{Introduction}

When a transformer answers a prompt, attention is how tokens talk to each
other~\cite{vaswani2017attention,elhage2021framework}. On a few well-studied behaviors (copying
a repeated pattern with induction heads~\cite{olsson2022induction}, or picking the right
name in a sentence with IOI~\cite{wang2023ioi}), that talk has been mapped into
sparse circuits. The catch is cost: the usual tools (patching, ACDC-style
search, attribution graphs) find those circuits by intervening again and
again~\cite{conmy2023acdc,heimersheim2024patching,ferrando2024ifr,ameisen2025circuittracing}.

So there is a gap. Either we trust raw attention as explanation (too weak), or
we pay for a full offline search (too heavy). We ask a middle question:

\begin{quote}
\emph{If we build a routing graph from a single forward pass and keep only a
small subgraph aimed at the answer, does that subgraph beat chance under
ablation?}
\end{quote}

If the answer is no, the idea is dead. If yes, we get a live sketch that is fast to
compute and at least partly causal, without pretending to replace heavier
methods.

That is the paper. Section~\ref{sec:method} walks through the recipe.
Section~\ref{sec:results} shows where it works, where it is incomplete, and
what breaks it.

\section{What we do}
\label{sec:method}

\begin{figure}[t]
  \centering
  \includegraphics[width=\linewidth]{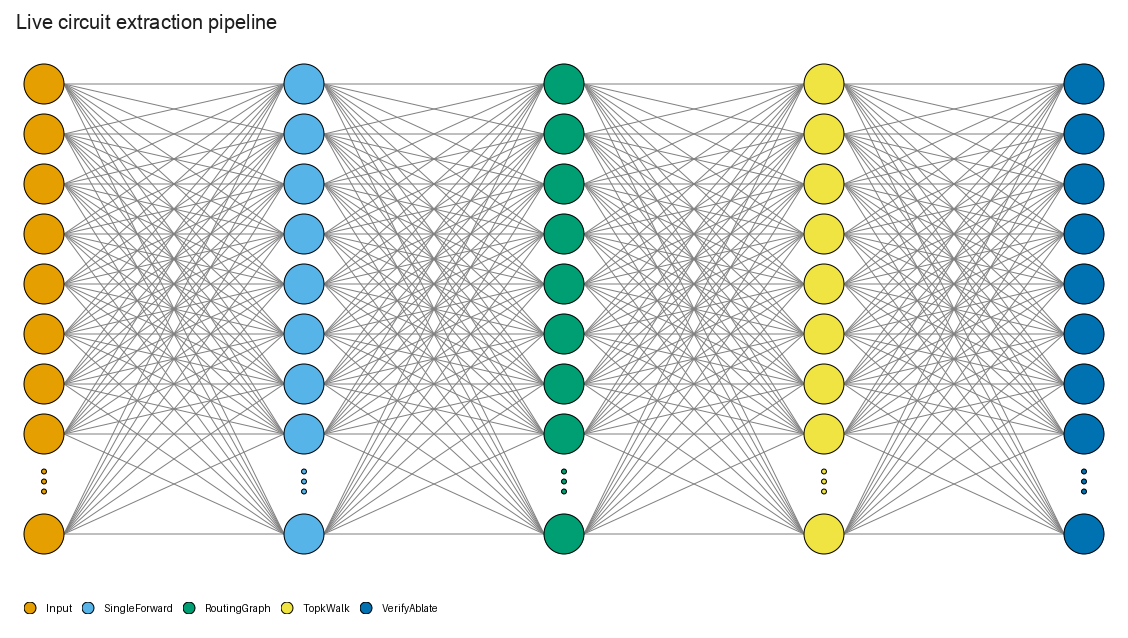}
  \caption{The idea in four steps: run the model once, read attention as a
  routing graph, walk backward from the answer to keep a small edge set \(S\),
  then check \(S\) against random edges of the same size.}
  \label{fig:pipeline}
\end{figure}

The pipeline is short on purpose (Figure~\ref{fig:pipeline}).

\paragraph{Read attention as a graph.}
Each token position is a node. An edge means: at some layer and head, this
query token attended to that key token. We keep head identity (induction needs
it), ignore the BOS sink, and skip self-loops. We score each edge by how hard
it attends times how large the value vector is, roughly ``how much was
written along this route'':
\[
s = A[\ell,h,i,j] \cdot \bigl\|V[\ell,h,j]\bigr\|_2 .
\]
Here \(V[\ell,h,j]\) is the pre-\(W_O\) value vector. That product sits next to a line of work that already warned against reading
raw attention alone. Kobayashi et al.~\cite{kobayashi2020attn} show that the
attention output depends on both the weight and the transformed value vector,
and analyze \(\|\alpha\,f(x)\|\) rather than \(\alpha\) by itself; our score is
a cheap edge-level cousin of that idea (weight times value norm), used only to
rank routes for extraction. Abnar and Zuidema~\cite{abnar2020rollout} go
further and compose attentions across depth (rollout / flow) because mixing
makes single-layer weights a poor token-importance probe. We take the shared
lesson that raw \(A\) is weak, but we do not roll out: we keep discrete
\((\ell,h,j\!\to\!i)\) edges, walk a few hops, and insist on a causal check.
Raw attention alone remains a backup score; the product works a bit better in
our ablations.

\paragraph{Keep a small subgraph.}
Start at the answer position \(t^\star\) (usually the last token). Take its
top-\(k\) incoming edges, then repeat a few hops backward from the tokens those
edges came from. Defaults are modest: \(k{=}15\), four hops. Call the kept set
\(S\).

\paragraph{Check that \(S\) matters.}
Turn those attention weights off (and renormalize what remains). Measure how
much the task score falls (target logit for induction, name logit-difference
for IOI). Do the same to ten random edge sets of size \(|S|\).
If \(S\) hurts clearly more than random, we call the sketch a pass.

That is the whole claim test. No dictionary training, no long search: just
routing, a short walk, and a fair control.

\section{What we find}
\label{sec:results}

We use GPT-2 Small as the main model, then replay induction on GPT-2 Medium and
Pythia-410M~\cite{biderman2023pythia}, and IOI on all three. Each cell uses
\(n{=}100\) clean prompts (single-token names/words after filtering). Same
recipe everywhere: product score, \(k{=}15\), four hops, ten random controls.
Besides means, we report a paired per-example gap
\(g_i=\Delta_S^{(i)}-\Delta_R^{(i)}\), a \(95\%\) bootstrap CI for the mean
gap, and a
one-sided Wilcoxon signed-rank test that \(\Delta_S>\Delta_R\).

\subsection{The sketch beats random}

\begin{figure}[t]
  \centering
  \includegraphics[width=0.92\linewidth]{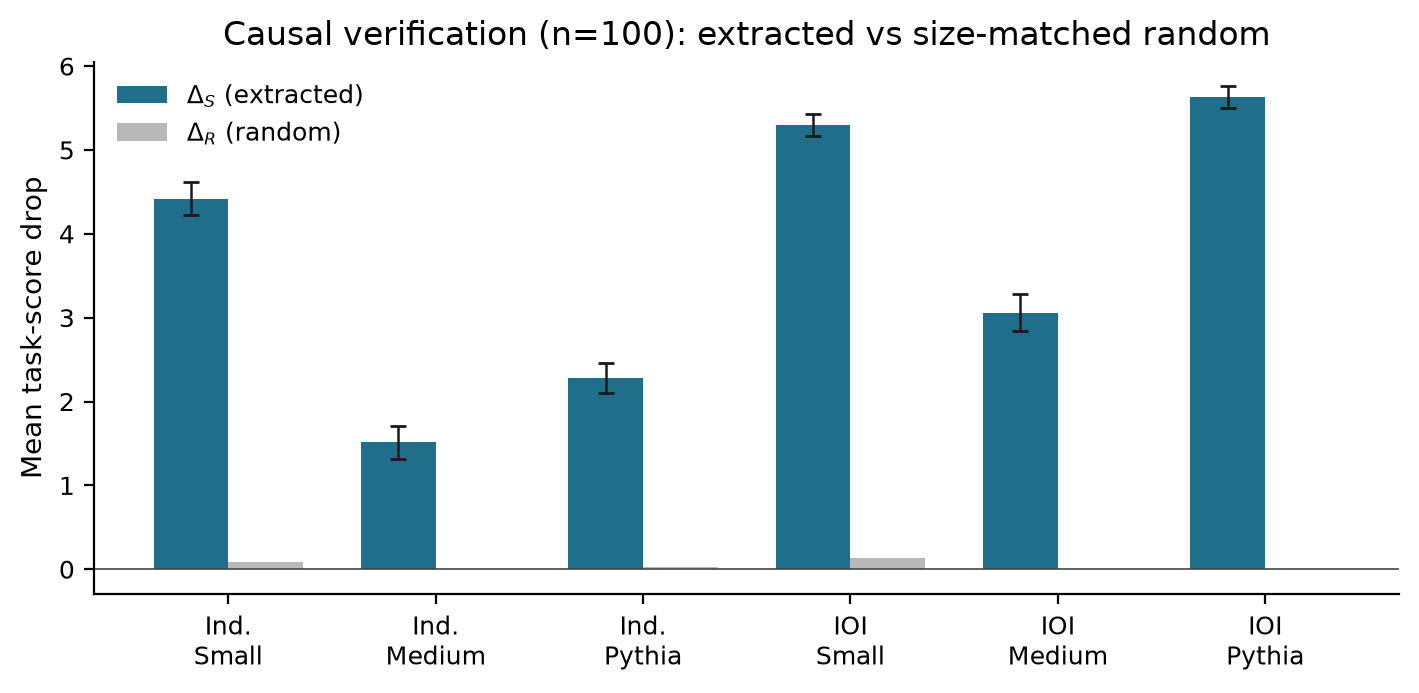}
  \caption{How much the task score falls when we ablate our edges (\(S\))
  versus random edges of the same size. Taller teal bars mean the sketch actually mattered. Error bars show \(\pm\) one standard error of the mean over prompts.}
  \label{fig:causal}
\end{figure}

Figure~\ref{fig:causal} and Table~\ref{tab:main} are the headline. Every cell
uses \(n{=}100\) and a paired gap test. On GPT-2 Small induction, turning off
\(S\) drops the target logit by about \(4.42\) versus \(0.09\) for random
(gap \(+4.33\); \(95\%\) bootstrap CI \([+3.95,+4.72]\); fails \(0/100\)).
Medium and Pythia induction keep positive gaps (\({+}1.51\), \({+}2.25\)) with
CIs that exclude zero, though more examples fail individually (\(19/100\),
\(11/100\)). IOI behaves the same way: Small and Pythia are essentially
clean (gaps \({+}5.16\) and \({+}5.63\); fails \(0/100\)), and Medium still clears a comfortable margin (gap \({+}3.05\); CI \([+2.63,+3.46]\); fails \(12/100\)).
In all six cells the Wilcoxon test that \(\Delta_S>\Delta_R\) is
\(p{<}10^{-10}\).

\begin{table}[t]
  \centering
    \caption{Causal verification at \(n{=}100\). \(\Delta_S\) and \(\Delta_R\) are
  mean score drops, \(\pm\) one standard deviation across prompts. Gap
  \(=\Delta_S-\Delta_R\), computed as the mean paired difference; columns are
  rounded independently, so the gap can differ from the displayed subtraction in
  the last digit. Every cell has a \(95\%\) bootstrap CI on the gap excluding
  \(0\) and Wilcoxon \(p{<}10^{-10}\); all six intervals are in the released
  artifacts.}
  \label{tab:main}
  \setlength{\tabcolsep}{5pt}
    \begin{tabular}{@{}llrrccc@{}}
    \toprule
    Task & Model & \(n\) & \(|S|\) & \(\Delta_S\) & \(\Delta_R\) & gap \\
    \midrule
        Induction & GPT-2 Small  & 100 &  71.0 & \(+4.42{\pm}1.97\) & \(+0.09{\pm}0.18\) & \textbf{+4.33} \\
    Induction & GPT-2 Medium & 100 &  61.2 & \(+1.51{\pm}1.93\) & \(\phantom{+}0.00{\pm}0.06\) & \textbf{+1.51} \\
    Induction & Pythia-410M  & 100 &  59.4 & \(+2.27{\pm}1.79\) & \(+0.03{\pm}0.11\) & \textbf{+2.25} \\
    IOI       & GPT-2 Small  & 100 & 195.0 & \(+5.30{\pm}1.31\) & \(+0.13{\pm}0.23\) & \textbf{+5.16} \\
    IOI       & GPT-2 Medium & 100 & 192.0 & \(+3.06{\pm}2.15\) & \(+0.01{\pm}0.09\) & \textbf{+3.05} \\
    IOI       & Pythia-410M  & 100 & 169.0 & \(+5.63{\pm}1.30\) & \(+0.01{\pm}0.09\) & \textbf{+5.63} \\
    \bottomrule
  \end{tabular}
\end{table}

\subsection{It is also cheap}

\begin{figure}[t]
  \centering
  \includegraphics[width=0.95\linewidth]{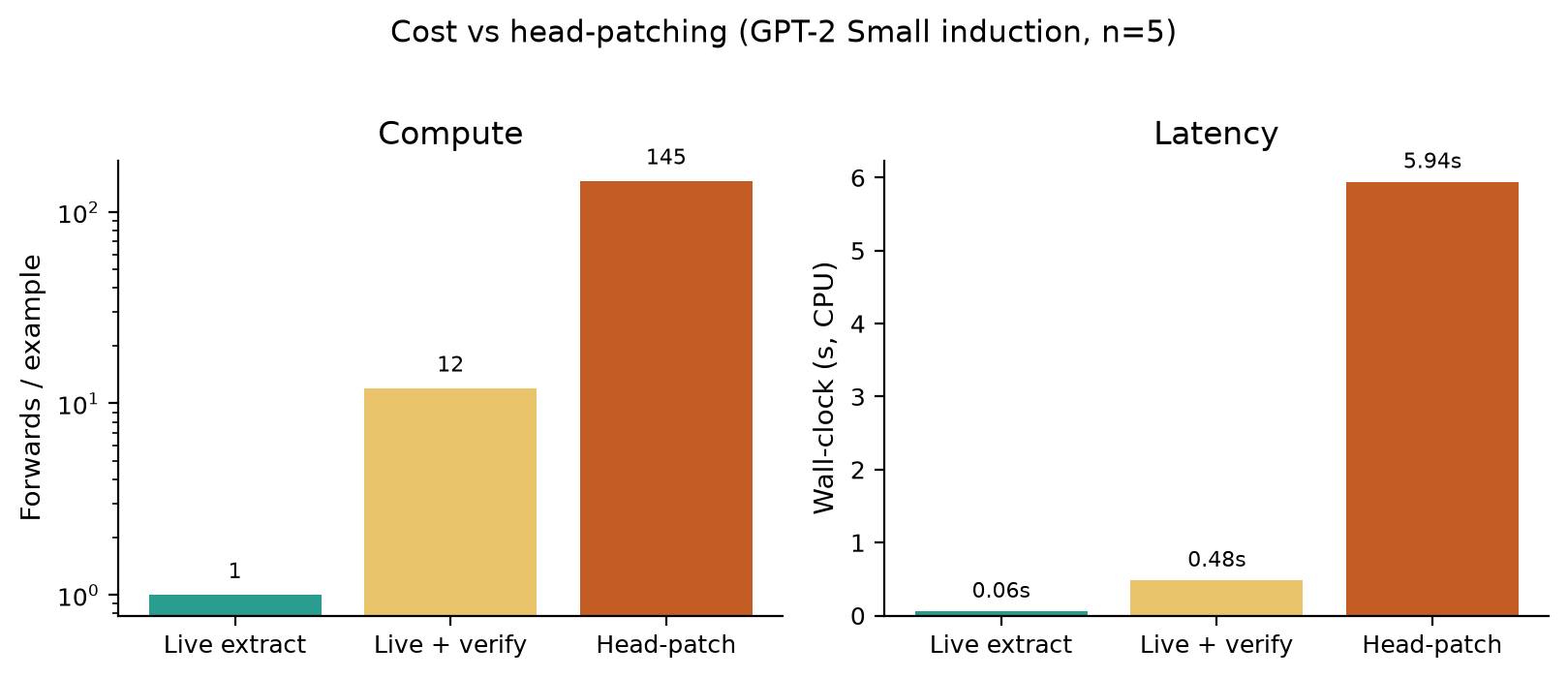}
  \caption{Why ``live'' matters: one forward to sketch, a few more to verify,
  versus a full head-by-head patch sweep.}
  \label{fig:cost}
\end{figure}

Figure~\ref{fig:cost} compares cost on Small induction. Building \(S\) is one
forward. Adding our verify step (ablate \(S\) plus ten random sets) is still
about a dozen forwards. Sweeping every head once needs \(145\). Wall-clock on
CPU tells the same story: roughly a hundred times faster to extract than to
head-patch. Full edge search would be slower still; the head sweep is already
a lower bound.

\subsection{It overlaps known heads, but not all of them}

\begin{figure}[t]
  \centering
  \includegraphics[width=\linewidth]{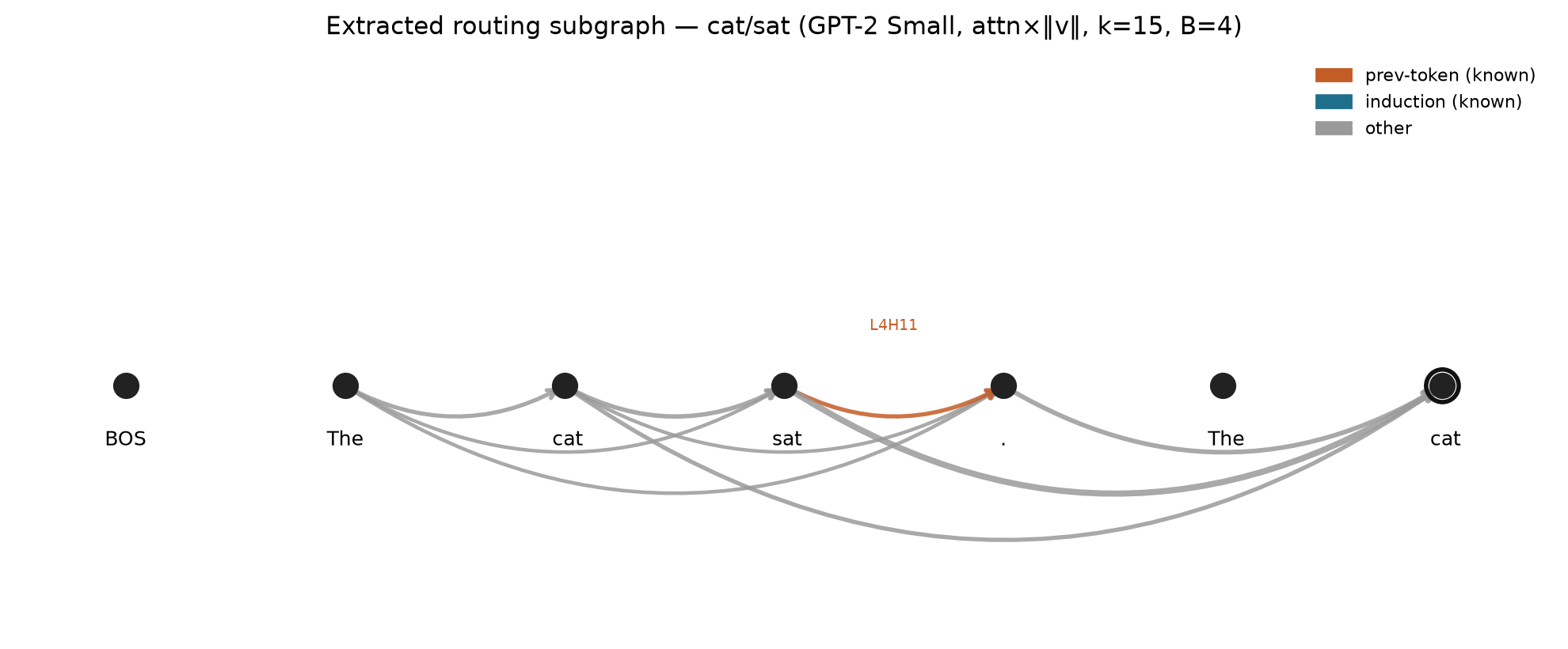}
  \caption{One extracted subgraph on
  \texttt{The cat sat. The cat}$\,\to\,$\texttt{sat}.
  Colored arcs mark known induction-related heads when they dominate a route.}
  \label{fig:subgraph}
\end{figure}

Figure~\ref{fig:subgraph} shows one induction example. The walk often picks up
famous previous-token and induction heads (L4H11, L5H5, L5H8 show up almost
always). Later named heads appear only sometimes. So a strong causal gap does
\emph{not} mean we recovered the textbook circuit end to end. It means we
caught enough of a working path that breaking it hurts.

\subsection{When it fails}

We count an example as a fail when \(\Delta_S\le\Delta_R\), and we also report
paired tests on \(g_i\) so the fail count is not doing all the persuasion.
Across the six \(n{=}100\) cells, fails concentrate on Medium induction
(\(19/100\)), Pythia induction (\(11/100\)), and Medium IOI (\(12/100\));
Small induction, Small IOI, and Pythia IOI are \(0/100\). Typical residual
modes: the walk latches onto a high-scoring distractor, the fixed hop budget
misses a longer path, or attention mass is not causation (hence the BOS ban
and the value-norm score).

\subsection{What we varied}

We poked the knobs on GPT-2 Small induction. Too small a walk (\(k{=}5\)) is
unreliable; \(k{\ge}10\) is fine, and our default \(k{=}15\) is in the stable
range. An overly harsh score threshold throws away the signal. Using only late layers fails. Early layers alone recover most of the gap, which fits the induction story: the routes that matter form early. The product score ranks routes better than raw attention, though the margin is small enough that either would support the main result. Changing the prompt template or padding a
few filler sentences does not kill the gap. In short: the result is not glued to one brittle setting, but it does need enough of the graph and the right layers. Per-setting gaps are in the released artifacts.

\section{What we are not saying}

This is a routing sketch with a causal check, not a full discovery engine. We
do not name features, train sparse autoencoders, or claim every edge in \(S\)
is necessary or monosemantic~\cite{marks2025sfc,ameisen2025circuittracing}.
We also do not replace attention rollout or a full norm-based attention
analysis~\cite{abnar2020rollout,kobayashi2020attn}: those target token-level
importance under mixing or a finer decomposition of \(\alpha f(x)\). Our score
is a one-forward ranking signal for candidate routes; the claim rests on the
ablation gap, not on the score being a complete attribution method. Heavier
circuit tools remain right when you need completeness
\cite{conmy2023acdc,ferrando2024ifr}. The wedge is simpler: attention already
draws a map; a short walk on that map is often enough to find something that
matters, fast.

We also stayed on small open models and two synthetic tasks. Transfer to larger
models or messier behavior is open. Verification zeros attention and renormalizes, and our control samples edges uniformly, so \(S\) removes far more attention mass than a random set of the same size does. Part of each gap is therefore a mass effect rather than a routing effect, and a mass-matched control would shrink it by an amount we have not measured. Other intervention styles might disagree. And \(|S|\) grows with
prompt length, so fixed \(k\) and hop depth will eventually under-cover.

\section{Conclusion}

Attention already tells you where information tried to go. We turn that into a
small subgraph aimed at the answer, then ask a blunt question: does breaking
those routes hurt more than breaking random ones? On induction and IOI, yes:
across a few models, at a fraction of the cost of head-patching, with honest
partial recovery and a short list of failure modes. The useful object is the
sketch plus the check, not a claim that the atlas is finished.

Code, configs, and run artifacts for the experiments in this paper are public
at \url{https://github.com/Aquinf03/live-circuit-routing}.

\section*{Acknowledgments}
This work was produced at Aquin Labs. GPT-2 checkpoints are released by OpenAI
under their model terms; Pythia by EleutherAI. Accept applicable licenses before
downloading weights.

\bibliographystyle{plain}
\bibliography{refs}

\end{document}